\documentclass{article}

    \PassOptionsToPackage{numbers, compress}{natbib}
    \usepackage[preprint]{neurips_2025}

\usepackage[utf8]{inputenc} % allow utf-8 input
\usepackage[T1]{fontenc}    % use 8-bit T1 fonts
\usepackage{hyperref}       % hyperlinks
\usepackage{url}            % simple URL typesetting
\usepackage{booktabs}       % professional-quality tables
\usepackage{amsfonts}       % blackboard math symbols
\usepackage{nicefrac}       % compact symbols for 1/2, etc.
\usepackage{microtype}      % microtypography
\usepackage{xcolor}         % colors
\usepackage{graphicx}
\usepackage{amsmath}
\usepackage{amssymb}
\usepackage{makecell}
\usepackage[shortlabels]{enumitem}
\usepackage[linesnumbered,ruled,vlined]{algorithm2e}
\SetKwInOut{Input}{Input}
\SetKwInOut{Output}{Output}
\usepackage{newtxtext,newtxmath}

\newcommand{\algo}{\textsc{HierRouter}}

\title{\algo: Coordinated Routing of Specialized Large Language Models via Reinforcement Learning} 

\author{%
  David S.~Hippocampus\thanks{Use footnote for providing further information
    about author (webpage, alternative address)---\emph{not} for acknowledging
    funding agencies.} \\
  Department of Computer Science\\
  Cranberry-Lemon University\\
  Pittsburgh, PA 15213 \\
  \texttt{hippo@cs.cranberry-lemon.edu} \\
}

\author{%
Nikunj Gupta$^{1}$\thanks{Equal contribution. Corresponding author: Nikunj Gupta \texttt{\{nikunj@usc.edu\}}} \quad
Bill Guo$^{1}$\footnotemark[1] \quad
Rajgopal Kannan$^{2}$ \quad
Viktor K. Prasanna$^{1}$ \\
$^{1}$ University of Southern California \\
$^{2}$ DEVCOM Army Research Office \\
\texttt{\{nikunj, billguo, prasanna\}@usc.edu} \\ 
\texttt{\{rajgopal.kannan.civ\}@army.mil}
}

\begin{document}

\maketitle

\begin{abstract}

Large Language Models (LLMs) deliver state-of-the-art performance across many tasks but impose high computational and memory costs, limiting their deployment in resource-constrained or real-time settings. To address this, we propose \algo, a hierarchical routing approach that dynamically assembles inference pipelines from a pool of specialized, lightweight language models. Formulated as a finite-horizon Markov Decision Process (MDP), our approach trains a Proximal Policy Optimization (PPO)-based reinforcement learning agent to iteratively select which models to invoke at each stage of multi-hop inference. The agent conditions on the evolving context and accumulated cost to make context-aware routing decisions. Experiments with three open-source candidate LLMs across six benchmarks, including QA, code generation, and mathematical reasoning, show that \algo\ improves response quality by up to \textbf{2.4$\times$} compared to using individual models independently, while incurring only a minimal additional inference cost on average. These results highlight the promise of hierarchical routing for cost-efficient, high-performance LLM inference. All codes can be found here \url{https://github.com/Nikunj-Gupta/hierouter}. 

\end{abstract}

\section{Introduction} 
\label{sec:Introduction}
  
The ever-growing scale of large language models (LLMs) has led to escalating computation and memory demands during inference, posing significant challenges for real-time deployment and scalability. These costs are particularly prohibitive in latency-sensitive or resource-constrained settings, where the time and energy required to generate responses may outweigh the benefits of improved accuracy. While architectural techniques such as mixture-of-experts \cite{cai2024survey} and speculative decoding \cite{xia2024unlocking} have been proposed to reduce token-level compute, they are often tightly coupled to specific model internals and offer limited flexibility or interpretability. Alternatively, there is growing interest in higher-level coordination strategies that can operate on a pool of specialized models to maintain task performance while systematically reducing resource usage. 

In parallel with efforts to optimize inference efficiency, there has been a growing trend toward the development and deployment of smaller language models that are tailored for specific domains or task types. These models leverage architectural innovations and training efficiencies to maintain high accuracy while drastically reducing computational and memory costs. Recent studies have demonstrated that, when carefully specialized, smaller models can rival or even outperform large-scale models on tasks aligned with their training objectives \cite{lu2024small,wang2024comprehensive,hsieh2023distilling}. For example, compact models tuned for question answering \cite{taori2023alpaca,gichamba2024colbert}, code generation \cite{team2024codegemma,guo2024deepseek,abdin2024phi,gunasekar2023textbooks}, or mathematical reasoning \cite{shao2024deepseekmath,azerbayev2024llemma,mishra2022lila,ahn2024large} often outperform general-purpose LLMs of much larger size on those respective tasks. This performance gain is attributed to task-specific representation learning and focused data curation, which allows these models to generalize efficiently within narrower operational scopes. As a result, the AI community has begun exploring ways to coordinate collections of such specialized models through routing or ensembling mechanisms rather than relying solely on increasingly larger monolithic systems. 

\begin{figure}[t]
    \centering
    \includegraphics[width=0.85\linewidth]{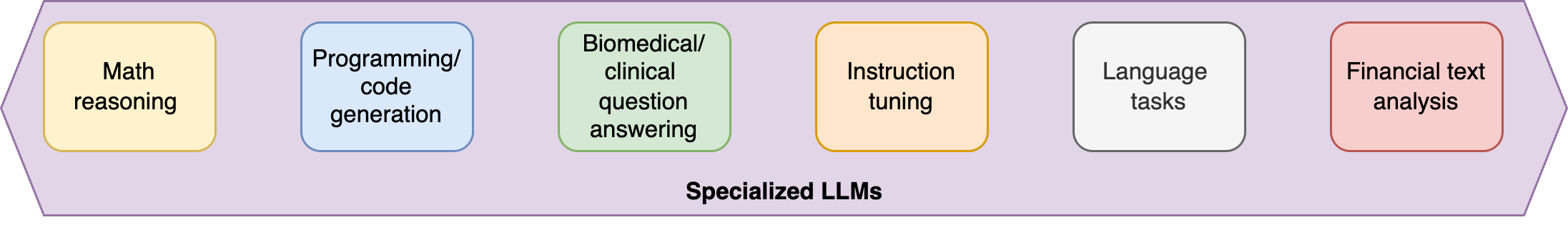}
    \caption{Specialized LLMs are smaller models fine-tuned for specific domains or task types, such as code generation, math reasoning, or biomedical QA. Unlike general-purpose LLMs, these models exhibit high efficiency and competitive accuracy on aligned tasks. By coordinating such models through routing rather than relying on monolithic architectures, systems like \algo\ can support adaptive, low-cost inference while preserving task performance.}
    \label{fig:specialized-llms}
\end{figure}

To effectively leverage the capabilities of smaller, specialized models, two broad coordination strategies have emerged: model assembly and routing. Assembly-based approaches aggregate outputs from multiple models via voting \cite{yang2024llm}, response fusion \cite{huang2024ensemble,wang2024fusing,jiang-etal-2023-llm}, or learned interpolation \cite{tay2023ul}. While these techniques enhance robustness by combining model strengths, they often require querying several models simultaneously, undermining efficiency and negating the benefits of using smaller models. For instance, frameworks like DeepEn \cite{huang2024ensemble} fuse outputs with task-specific transfer matrices but introduce substantial memory and runtime overhead. In contrast, routing-based approaches \cite{chen2024routerdc,lu-etal-2024-routing,ding2024hybrid} select a single model (or a small subset) per query to minimize redundancy. However, most rely on one-shot decisions made upfront, limiting adaptability for complex queries requiring iterative reasoning or dynamic cooperation. Cascading approaches \cite{chen2024frugalgpt} offer partial adaptivity, triggering stronger models when simpler ones fall short, but still depend on rigid heuristics and lack contextual flexibility. These limitations highlight the need for a flexible, multi-step routing framework that treats inference as a sequential decision process, enabling context-aware model selection, refinement of partial answers, and improved quality-cost tradeoffs through adaptive reasoning.

We propose \algo, a reinforcement learning (RL)-based hierarchical routing approach that dynamically orchestrates a pool of specialized language models in a multi-hop architecture. At each step of inference, a learned router selects a single model from the candidate pool to process the current query or intermediate context. The selected model generates a response, which is appended to the context and passed forward to the next routing stage. This process continues for a fixed number of hops, enabling compositional reasoning and progressive refinement of the response. Crucially, our RL-based router is trained to optimize a reward function that balances task performance against cumulative inference cost. Unlike prior approaches that rely on static ensembling, one-shot routing, or brittle reward alignment, our router learns a context-aware policy that adapts its decisions based on both semantic progress and resource usage. By framing routing as a reinforcement learning task over a finite-horizon Markov Decision Process, we enable generalization to novel queries and dynamic model pools while ensuring robust model coordination over time. This structure captures model specialization and supports incremental inference, where earlier hops address simpler subproblems and later hops escalate complexity as needed. Compared to prior methods, \algo\ offers a more interpretable and cost-efficient coordination mechanism: it builds answers step-by-step, leverages diverse model strengths, and allocates compute resources only where necessary. This mirrors how human experts solve problems through gradual synthesis, consultation, and refinement, making \algo\ a more natural and effective strategy for LLM orchestration. 

\begin{itemize}
    \item We develop a novel hierarchical routing mechanism that iteratively selects specialized models at each hop to incrementally refine the response over a fixed inference horizon.
    
    \item Our PPO-based router is trained using a sparse reward that balances task accuracy against cumulative cost, enabling flexible adaptation to both semantic content and compute budgets.

    \item We conduct extensive experiments across six diverse benchmarks, including question answering (MMLU, ARC), code generation (MBPP), and mathematical reasoning (GSM8K, MATH), and show that \algo\ improves average quality score by up to \textbf{107\%} compared to the best small model, while maintaining reasonably close inference cost compared to individual small (Qwen2.5-Coder-3B, DeepSeek-R1-Distill-Qwen-1.5B, Phi-3.5-mini) and bigger LLMs (Llama-3.1-8B, Qwen2.5-14B).
\end{itemize}

\section{Related works}
\label{sec:Related}
 
\paragraph{Inference Cost and Efficiency in LLMs.} While LLMs demonstrate impressive performance across a wide array of tasks, their substantial inference costs, measured in latency, memory, and energy consumption, pose significant challenges for deployment, especially in real-time or resource-constrained settings. Several efforts have focused on reducing these costs through model compression techniques such as pruning, quantization, and knowledge distillation~\citep{ma2023llm,egashira2024exploiting,liu2024ddk}. Others have explored dynamic inference strategies, including early exiting~\citep{elhoushi2024layerskip}, input-dependent routing~\citep{chen2024routerdc}, and expert selection in mixture-of-experts architectures~\citep{cai2024survey}. Despite these advances, balancing task performance with compute efficiency remains a critical challenge, particularly in scenarios involving diverse or unseen inputs where static optimization methods fall short. 

\paragraph{Rise of Small Specialized Models.} The growing ecosystem of task-specialized language models reflects a broader trend toward modularity and efficiency in LLM deployment. Recent works have introduced lightweight models fine-tuned for specific domains such as programming~\citep{team2024codegemma,guo2024deepseek,abdin2024phi,gunasekar2023textbooks}, math reasoning~\citep{shao2024deepseekmath,azerbayev2024llemma,mishra2022lila,ahn2024large}, or retrieval-augmented QA~\citep{asai2023self,liu2024can}. These models often outperform larger, general-purpose models in their respective domains, while requiring significantly fewer resources. This has motivated research into systems that intelligently coordinate such models. Our work builds on this paradigm by proposing a reinforcement learning-based router that learns to compose and coordinate specialized small LLMs in a cost-sensitive, multi-hop inference pipeline. 

\paragraph{Model Coordination Strategies.} Coordinating multiple models for improved inference has been explored through both ensembling and routing paradigms. Ensembling-based approaches typically aggregate outputs from multiple models, either via majority voting, confidence weighting, or learned fusion, to boost accuracy or robustness~\citep{yang2024llm,huang2024ensemble,wang2024fusing,jiang-etal-2023-llm,tay2023ul}. While effective, ensembles are computationally expensive due to redundant model evaluations. In contrast, routing-based approaches aim to select a single model (or a subset) per input, reducing cost while maintaining accuracy~\citep{chen2024routerdc,lu-etal-2024-routing,ding2024hybrid,ong2025routellm}. These methods often rely on heuristics or input similarity measures to assign queries to appropriate models. Moreover, most routing strategies are shallow (one-shot) and lack mechanisms for iterative refinement or composition, limiting their expressivity and adaptability in complex reasoning tasks. 

% \paragraph{Reward-Based Routing and Its Limitations.} Recent works have proposed learning routing policies using reward-based supervision, often leveraging response-level feedback such as answer correctness or external reward model scores~\citep{mialon2023augmented, zhou2023llm}. While these methods enable task-aware model selection, they face challenges with sparse or delayed rewards, unstable reward alignment, and lack of generalization across domains. Moreover, many reward-based routers operate in a single-hop fashion, making decisions in isolation without leveraging feedback from intermediate model responses. This limits their capacity for multi-step reasoning or correction. Our approach addresses these limitations by casting model routing as a sequential decision-making problem and using reinforcement learning to train a multi-hop router that conditions on both evolving context and cumulative inference cost. 

\paragraph{LLM Cascading.} Cascading strategies execute LLMs in sequence, typically escalating queries from smaller to larger models based on fixed heuristics or confidence thresholds~\citep{chen2024frugalgpt,zhang2024ecoassistant,varshney-baral-2022-model,aggarwal2024automix}. These systems aim to reduce compute by resolving easy queries early, but are often rule-based, brittle to distribution shifts, and difficult to adapt to heterogeneous tasks. Moreover, model composition in cascading pipelines is typically overwrite-based: later models discard earlier outputs and reprocess the query from scratch, limiting opportunities for incremental reasoning. 
% Unlike trainable approaches, these methods lack formal supervision or learning signals, which restricts their flexibility and generalization. 
In contrast, our work introduces a trainable routing agent that conditions model selection on evolving context and accumulated cost, supporting multi-hop composition and adaptive refinement across inference steps.

\paragraph{Policy Optimization in RL.} Policy gradient methods form the foundation of modern RL for decision-making tasks. Among them, Proximal Policy Optimization (PPO) ~\citep{schulman2017proximal} has emerged as a robust and sample-efficient algorithm widely used in natural language and control domains. PPO improves training stability by clipping updates to the policy objective, making it well-suited for sparse reward settings such as in this paper. In this work, we leverage PPO to train our routing agent. 

% \paragraph{Reinforcement Learning in LLM Routing.} Reinforcement learning has emerged as a promising framework for optimizing model usage in large-scale inference pipelines. Several recent works apply RL to train routing policies that decide which LLM to invoke based on context or task difficulty~\citep{}. These methods typically define reward signals based on output quality, latency, or compute cost, and use policy gradient methods to learn adaptive strategies. However, most existing RL-based routers are limited to one-shot decisions, lack temporal credit assignment, or assume a fixed set of routing targets without feedback. Our work builds on this foundation but introduces a hierarchical multi-hop setting, where an RL agent sequentially selects models over multiple stages of inference. By modeling routing as a finite-horizon MDP, we enable compositional reasoning, cost-aware exploration, and iterative refinement, capabilities that are largely missing from prior RL-based routing formulations. 

% \section{Preliminaries}
% \label{sec:Preliminaries}
% \input{sections/03_preliminaries}

\section{Methodology}
\label{sec:Methodology}

\begin{figure}[t]
    \centering
    \includegraphics[width=0.75\linewidth]{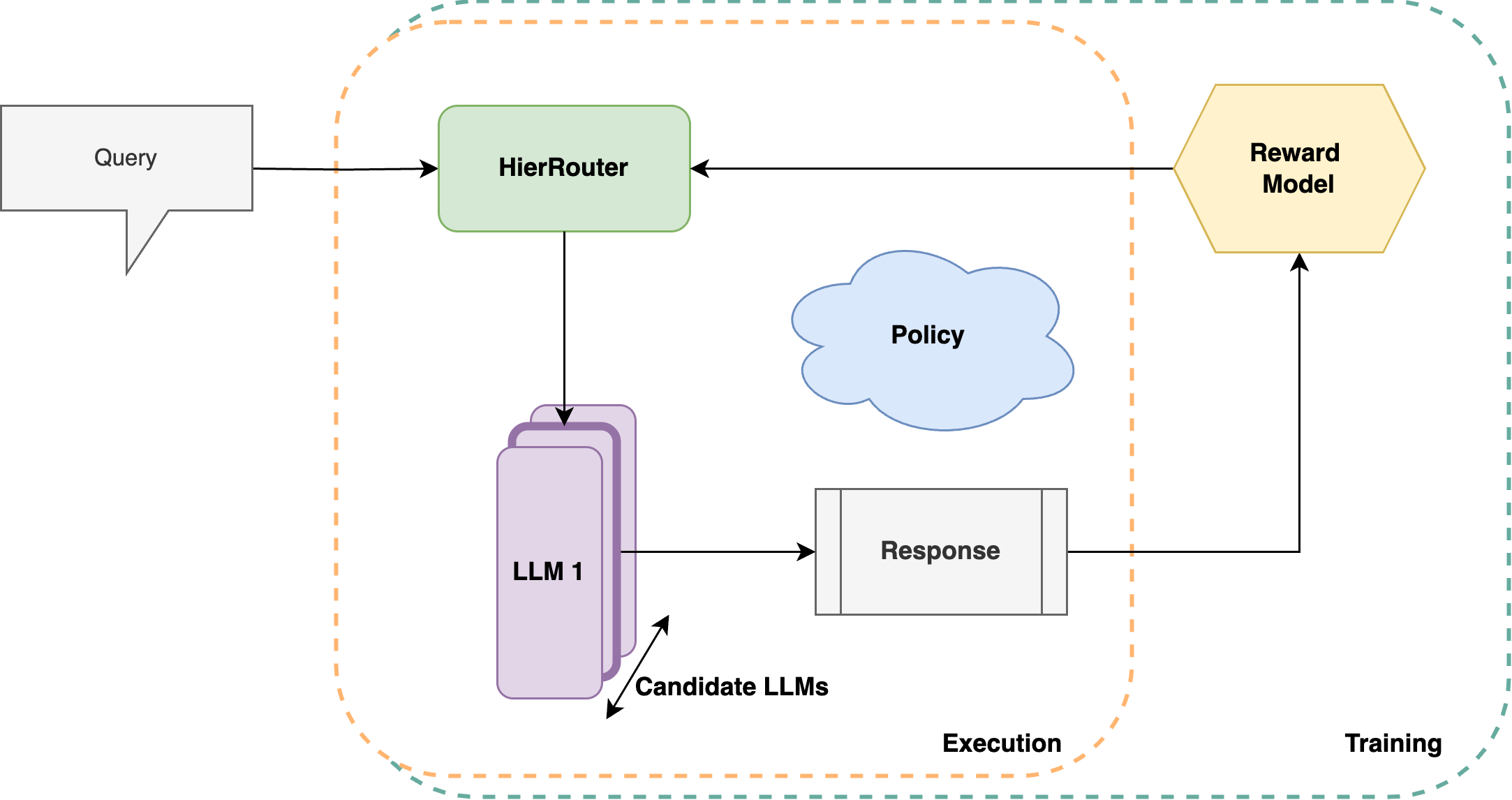}
    % \caption{Overview of the \algo\ architecture. The system operates in a multi-hop fashion, where at each step, a learned PPO-based router selects a subset of specialized language models (e.g., \texttt{model1}, \texttt{model2}, …, \texttt{model3}) to respond to the current query or intermediate representation. The outputs from the selected models are then passed to a lightweight fusion module that aggregates them into a unified response. This fused response is fed back into the next router layer, which iteratively refines the answer by selecting new model subsets. The process continues for a fixed number of hops or until convergence criteria are met. By deferring complex decisions and progressively accumulating information through model cooperation, the system achieves better performance-cost trade-offs compared to one-shot routing or static ensembling. This design also supports compositional reasoning, where earlier hops handle simpler aspects of the query while later hops refine or escalate complexity as needed.}
    \caption{Overview of the \algo\ architecture. At each of the \(L\) inference stages, a PPO-based (RL) router selects one model from a pool of specialized LLMs to respond to the current context. The model's output is appended to the evolving context, which is passed to the next router layer. This multi-hop process enables compositional reasoning: earlier hops handle simpler sub-tasks, while later hops refine the output. The routing policy is trained to optimize response quality under inference cost constraints, balancing accuracy and efficiency through sequential model coordination.}
    \label{fig:hierrouter}
\end{figure}

In this section, we describe our approach, \algo, for hierarchically and adaptively selecting from a pool of specialized LLMs. The goal is to learn a cost-aware policy that adaptively selects models at each inference step to balance output quality and compute efficiency. Figure~\ref{fig:hierrouter} illustrates the high-level architecture of \algo. The methodology consists of three main components: (i) the formalization of the routing process as a Markov Decision Process, (ii) the design of the router policy network for model selection, and (iii) the PPO-based training procedure for learning cost-sensitive, context-aware routing strategies. 

\subsection{MDP Formulation}\label{sec:mdp_formulation}

To jointly optimize response quality and computational efficiency in the context of hierarchical LLM routing, we formalize the problem as a finite-horizon Markov Decision Process (MDP):
\[
  \mathbb{M} \;=\; \langle \mathcal{S}, \mathcal{A}, P, R, \gamma, L \rangle,
\]
where the agent operates over a structured decision space to construct a compositional inference pipeline. At each decision step, the agent selects one model from a fixed pool of \(M\) specialized LLMs. The process unfolds over a fixed number of \(L\) stages, reflecting the bounded depth of the routing hierarchy. This formulation captures the essence of dynamic inference in multi-model systems, where partial outputs at each stage serve as intermediate context for future decisions. The agent must reason over evolving semantic representations while balancing two competing objectives: (i) maximizing the quality of the final output, as judged by an external or learned evaluator, and (ii) minimizing the total computational cost, typically measured via token usage or latency. % By casting the routing problem within an MDP framework, we enable principled learning of adaptive policies that make context-aware, cost-sensitive model selection decisions in a sequential manner. 

\paragraph{State space.} At each decision stage \(t\), the agent observes a state \(s_t = (c_t, \ell_t, C_t)\), which compactly encodes all information necessary for selecting the next routing action. The component \(c_t\) denotes the context history - a growing textual sequence formed by concatenating the original user query with the responses generated by each selected model up to the current point. This evolving context serves both as the input to subsequent models and as a representation of the semantic progress made so far. The second component, \(\ell_t \in \{0, \dots, L-1\}\), indicates the current depth in the decision process, reflecting how many stages of the routing hierarchy have been executed. This acts as a surrogate for the residual budget or inference horizon. Finally, \(C_t \in \mathbb{R}_{\ge 0}\) captures the cumulative inference cost incurred so far, such as token-FLOPs or latency-normalized resource usage, which directly influences the reward signal. Together, these three components define a state space that balances semantic reasoning and budget awareness. The agent must learn to interpret \(c_t\) to assess what aspects of the query have been resolved, use \(\ell_t\) to track the decision horizon, and consider \(C_t\) to control computational efficiency. % This joint representation is crucial for enabling cost-sensitive routing policies that can adaptively terminate or refine the inference pipeline based on evolving task needs and remaining resources. 

\paragraph{Action space.} At each decision point \(t\), the agent selects an action \(a_t \in \mathcal{A}\), which corresponds to choosing a model \(m_t \in \{0, \dots, M - 1\}\) from the pool of \(M\) available language models to invoke next. Since the routing process proceeds for a fixed number of stages \(L\), the action space is simply of size \(|\mathcal{A}| = M\). This formulation enables the agent to focus on model selection at each hop without needing to reason about termination decisions. Crucially, by learning context-dependent model choices across stages, the agent can implement non-myopic routing strategies that optimize for long-term trade-offs between inference cost and final answer quality, leveraging the diverse capabilities of the underlying model ensemble. 

\paragraph{Transition dynamics.} The environment evolves deterministically according to the agent’s action and the behavior of black-box components \footnote{By "black-box functions" we mean the policy interacts with them only via inputs/outputs, without access to their internal mechanisms. This allows flexible integration of heterogeneous models and cost estimators.} for text generation and cost estimation (algorithm included in Appendix~\ref{alg:env-step}). We define the \texttt{Gen} function as a black-box model inference interface that takes the current context \(c_t\) and a selected model \(m_t\), and returns a textual response \(r_t = \texttt{Gen}(m_t, c_t)\). 
% Internally, this function may invoke a HuggingFace or OpenAI model via prompt completion, using frozen or pre-configured decoding settings such as temperature, top-\(p\), or max tokens. 
Each model has a predefined or dynamically inferred inference schema, including instruction formatting and tokenizer handling, which is abstracted from the agent. The \texttt{Cost} function computes the normalized inference cost \(\delta_t = \texttt{Cost}(m_t, c_t)\), using token-level statistics and per-model pricing metadata. For each call, we approximate the cost as:
\[
\delta_t = \text{base\_rate}(m_t) \cdot (\texttt{num\_tokens\_in} + \texttt{num\_tokens\_out}),
\]
where the base rate is obtained from either a user-supplied configuration file or estimated heuristically based on model size (e.g., 3B, 7B, 13B) when exact pricing is unavailable. The environment ensures cost units are normalized (e.g., USD/token or FLOPs/token), making the resulting penalty comparable across diverse architectures. This abstraction enables the router to reason over both semantic and budgetary signals in a unified reward formulation. 

Specifically, upon selecting action \(a_t = m_t\) at stage \(t\), the agent queries model \(m_t\) with the current context \(c_t\). This results in a new textual response \(r_t = \texttt{Gen}(m_t, c_t)\), generated by the black-box function \texttt{Gen}, and an associated computational cost \(\delta_t = \texttt{Cost}(m_t, c_t)\), computed by a black-box estimator \texttt{Cost}. The cost function typically reflects a proxy for inference effort such as normalized token-level FLOPs, model-specific latency, or memory usage. The environment then transitions to the next state \(s_{t+1} = (c_{t+1}, \ell_{t+1}, C_{t+1})\) where: $c_{t+1} = c_t \,\Vert\, r_t, \quad \ell_{t+1} = \ell_t + 1, \quad C_{t+1} = C_t + \delta_t.$ That is, the context is extended via concatenation with the new response \(r_t\), the layer index is incremented by one to reflect deeper traversal into the inference pipeline, and the cumulative cost is updated to include the additional compute incurred by \(m_t\). An episode ends when the routing depth reaches its predefined maximum \(\ell_{t+1} = L\). These transitions enforce a monotonic structure: context and cost only grow, and depth strictly increases. Conceptually, this models a progressive assembly of the final output, where each model’s contribution builds upon the prior state and incurs additive inference cost. By encoding both semantic and resource dynamics in state transitions, the agent can learn to balance output refinement against compute efficiency over time. 

\paragraph{Reward function.}  We define a sparse terminal reward that captures the trade-off between response quality and inference cost. The agent receives no intermediate rewards and is evaluated only at the final step \(T = L\), corresponding to the last stage of the multi-hop routing process. The reward function is given by:
\[
  R(s_T, a_T) = Q(c_T) - \alpha C_T,
\]
where \(Q(c_T) \in [0,1]\) denotes the quality of the final response as assigned by an external evaluator (e.g., a reward model, task metric, or oracle label), and \(C_T \in \mathbb{R}_{\ge 0}\) represents the total cumulative cost incurred over the trajectory. The cost penalty coefficient \(\alpha > 0\) controls the balance between performance and efficiency. This formulation encourages policies that produce high-quality answers while minimizing inference cost, and supports deployment in settings constrained by latency, memory, or energy budgets. The reward is agnostic to the specific form of the evaluator \(Q(\cdot)\), allowing flexibility in incorporating ground-truth metrics or learned preference models across different tasks.

\begin{algorithm}[t]
\caption{\algo: Hierarchical Multi-Hop Routing with PPO}
\label{alg:hierrouter}
\SetKwInOut{Input}{Input}
\SetKwInOut{Output}{Output}
\Input{Pool of models $\mathcal{M} = \{m_1, \dots, m_M\}$, max hops $L$, reward weight $\alpha$}
\Output{Trained policy $\pi_\theta(a|s)$ and value function $V_\phi(s)$}

\BlankLine
\ForEach{training episode}{
    Initialize context $c_0$, cost $C_0 \gets 0$, depth $\ell_0 \gets 0$\;
    \For{$t = 0$ \KwTo $L-1$}{
        Construct state $s_t = (c_t, \ell_t, C_t)$\;
        Sample model $m_t \sim \pi_\theta(\cdot \mid s_t)$\;
        Query model: $r_t \gets \texttt{Gen}(m_t, c_t)$\;
        Compute cost: $\delta_t \gets \texttt{Cost}(m_t, c_t)$\;
        Update context: $c_{t+1} \gets c_t \Vert r_t$\;
        Update cost: $C_{t+1} \gets C_t + \delta_t$\;
        Update depth: $\ell_{t+1} \gets \ell_t + 1$\;
        Store transition $(s_t, m_t, \delta_t, r_t)$\;
    }
    Evaluate final output $Q(c_L) \in [0,1]$\;
    Compute terminal reward: $R \gets Q(c_L) - \alpha C_L$\;
    Compute advantage estimates using GAE \cite{schulman2015high}\;
    Update $\pi_\theta$ and $V_\phi$ using PPO loss \cite{schulman2017proximal}\;
}
\end{algorithm} 

\subsection{\algo: Multi-Hop Model Routing}
% \paragraph{Learning objective.} 
We adopt an episodic learning framework in which the agent receives reward only at the terminal timestep. 
% Accordingly, we set the discount factor \(\gamma\) to reflect an undiscounted return, as intermediate steps do not emit any reward signal. 
Each episode consists of exactly \(L\) decision steps, corresponding to the fixed depth of the hierarchical routing procedure. This encourages the agent to be judicious in model selection at each step, as all hops contribute to the final reward. To learn routing behaviors that optimize the trade-off between response quality and computational cost, we train a parameterized stochastic policy \(\pi_\theta(a \mid s)\), along with a value function baseline \(V_\phi(s)\), using Proximal Policy Optimisation (PPO)~\cite{schulman2017proximal} (where $\theta$ and $\phi$ represent the parameters of the policy and value networks, respectively). The objective is to maximize the expected episodic return:
\[
  J(\theta) = \mathbb{E}_{\pi_\theta} \left[\sum_{t=0}^{T} R(s_t, a_t)\right],
\]
where trajectories \((s_0, a_0, \dots, s_{T}, a_{T})\) are sampled under the current policy \(\pi_\theta\). Since rewards are sparse and delayed until termination, policy gradients must propagate credit through sequences of compute-augmenting decisions. This formulation enables the agent to discover adaptive routing schemes that leverage the heterogeneous strengths of the model ensemble, while automatically adapting to task difficulty and remaining within stringent compute budgets (see Algorithm~\ref{alg:hierrouter}). 

We implement the routing agent as a shared neural network that jointly parameterizes the policy \(\pi_\theta(a \mid s)\) and value function \(V_\phi(s)\). For each input state \(s_t = (c_t, \ell_t, C_t)\), we encode \(c_t\) using a frozen Sentence-Transformer \cite{song2020mpnet}, and represent \(\ell_t\) via a learned stage embedding. These are concatenated to form a joint state representation. This representation is fed into a two-layer MLP to produce (i) action logits over the model pool, and (ii) a scalar value estimate. 

At inference time the router executes a \emph{multi‑hop} decision process governed by the learned policy \(\pi_{\theta}(a_t \mid s_t)\). 
Starting from the initial state \(s_0=(c_0,\ell_0{=}0,C_0{=}0)\), the agent iteratively selects an action
\(
  a_t = m_t \in \{0,\dots,M\!-\!1\},
\)
where \(m_t\) indexes the next LLM to query.  
The chosen model receives the current context \(c_t\) and produces a reply
\(r_t = \texttt{Gen}(m_t,c_t)\).  
We append this reply to yield the updated context  
\(
  c_{t+1} = c_t \,\Vert\, r_t,
\)  
increment the stage counter \(\ell_{t+1}=\ell_t+1\), and accumulate cost \(C_{t+1}=C_t+\texttt{Cost}(m_t,c_t)\).
% \paragraph{Iterative feedback loop.} 
As the context is continually enriched by earlier replies, deeper hops operate on a progressively more informative input.  
This feedback enables \emph{refinement}: specialised models correct or elaborate on partial answers produced by shallower models, while the router monitors accumulated cost \(C_t\) to decide whether further improvement justifies additional computation.  
Figure~\ref{fig:hierrouter} visualises the resulting decision stack. The episode concludes after \(L\) hops, at which point the reward
\(R(s_T, a_T)\) defined in Section~\ref{sec:mdp_formulation}
is issued (only when training). The sparse, final‑state reward encourages the policy to discover routing strategies that attain high‑quality answers with minimal compute by selecting efficient sequences of model invocations.

\section{Experiments}
\label{sec:Experiments}

\paragraph{Candidate LLMs.} The router operates over a curated pool of three specialized instruction-tuned LLMs, each offering complementary strengths: \emph{Qwen2.5-Coder-3B} \cite{hui2024qwen2}, tailored for efficient code understanding and generation; \emph{DeepSeek-R1-Distill-Qwen-1.5B} \cite{guo2025deepseek}, a general-purpose model distilled for lightweight reasoning (math and programming); and  \emph{Phi-3.5-mini} \cite{abdin2024phi}, optimized for compact instruction following and broad generalization. These serve as the decision space for the hierarchical policy during both training and inference. To contextualize the performance and cost trade-offs, we additionally evaluate the router against two larger models: \emph{Llama-3.1-8B} \cite{grattafiori2024llama} and \emph{Qwen2.5-14B} \cite{qwen2.5}. These are not part of the routing pool but are included in evaluation to highlight the benefits of routing over smaller, specialized models versus direct usage of larger monolithic ones. All models are queried under a consistent configuration (see appendix). 

\paragraph{Dataset Construction.} We train the router on a curated set of six tasks spanning math reasoning, program synthesis, and general QA. Each dataset is capped at 300 examples (randomly sampled) to ensure balanced task coverage while keeping the overall training lightweight. This setup is sufficient for learning a performant routing policy due to the shared representation backbone and task-agnostic reward function. The dataset pool includes: \emph{GSM8K} \cite{cobbe2021training}, \emph{MATH} \cite{hendrycksmath2021}, \emph{MBPP} \cite{austin2021program}, and \emph{ARC} \cite{clark2018think} (easy and challenge), and \emph{MMLU} \cite{hendrycks2020measuring}. MMLU queries are reformatted to use consistent multiple-choice labels (A/B/C/D). We use a random $\sim$70\%-30\% train-test split, ensuring consistent evaluation without overlap. Evaluation is conducted on the disjoint held-out test sets from the same task domains to ensure fair comparisons across all experiments, including static routing with small or large LLMs and our hierarchical router. While the experiments are run with relatively smaller subsets of the data points for demonstration, the framework is designed for seamless scalability to larger datasets and tasks. The reported metric values are interpreted comparatively to highlight trade-offs in quality and cost across strategies. 

\paragraph{Baselines.} To assess the effectiveness of our router, we compare it directly against each constituent model in isolation, both from the router's internal pool of smaller LLMs and a set of larger, higher-capacity LLMs. These comparisons reveal how the router's selective multi-hop coordination compares to running small or large models in a fixed, single-shot manner. Through these experiments, our focus is to demonstrate that: (i) a lightweight router trained over small, specialized models can match or surpass larger models in performance, and (ii) the same setup provides reasonable efficiency gains under identical conditions. This setup is necessary and sufficient to validate the router’s ability to dynamically trade off quality and cost across tasks of varying complexity. 

\paragraph{Implementation details.} The router is implemented as a two-layer feedforward neural network that jointly parameterizes both the policy and value function. At each decision step, the input state consists of three components: (i) an embedding of the evolving context \(c_t\), obtained using a frozen Sentence-Transformer encoder \cite{song2020mpnet}; (ii) a learned stage embedding representing the current routing depth \(\ell_t\); and (iii) a scalar indicating cumulative cost \(C_t\). These features are concatenated into a unified state vector and passed through a multilayer perceptron to produce model-selection logits and value predictions. Training is performed using PPO with generalized advantage estimation (GAE) \cite{schulman2015high}, using \(\gamma = 0.99\), bias-variance trade-off parameter \(\lambda = 0.95\), clip parameter 0.2, and entropy regularization coefficient 0.01. At the end of each routing episode the agent receives a terminal reward \( R = Q(c_T) - \alpha C_T \). The task-specific quality score \(Q(c_T)\) is computed by a ground-truth evaluator, which compares the final output to reference answers using an appropriate metric for the task. The evaluator includes normalization routines (e.g., punctuation stripping, case folding) and pattern-matching logic for robust label extraction. This design ensures metric consistency across heterogeneous benchmarks while maintaining reward granularity sufficient for policy optimization. The total inference cost accumulated over the episode, \(C_T\), is normalized using static weights proportional to model size (i.e., parameter count), for fair comparison of compute efficiency across models of varying scale. The cost penalty \(\alpha = 0.005\) is a hyperparameter and was tuned empirically.

\section{Results}
\label{sec:Results}

We evaluate \algo\ across six diverse benchmark datasets: \textbf{MMLU}, \textbf{GSM8K}, \textbf{MBPP}, \textbf{MATH}, \textbf{ARC-Easy}, and \textbf{ARC-Challenge}, covering general knowledge, mathematical reasoning, and code generation. Table~\ref{tab:acc_comparison} reports the average test quality score per dataset for three model groups: (i) \textit{large LLMs} (Qwen2.5-14B and Llama-3.1-8B), (ii) \textit{candidate small LLMs} included in the routing pool (Qwen2.5-Coder, DeepSeek-R1-Distill-Qwen-1.5B, and Phi-3.5), and (iii) our proposed method, \textit{\algo}, which dynamically selects a model per query via a multi-hop routing policy. 

For all experiments, we report the \textit{quality score} (Table \ref{tab:acc_comparison}) as the token-level F1 score between the model-generated answer and the reference answer, consistent with common evaluation practices for open-ended QA and reasoning tasks. 
% All models are scored using token-level F1, providing a more robust and informative signal than exact match for open-ended tasks such as math and code generation. 
F1 provides a more robust and informative signal than an exact match, which can be too brittle in some cases (e.g., math derivations or code outputs with lexical variation). 
% combines both precision and recall, making it robust to partial correctness, particularly useful in tasks where exact match is too brittle (e.g., math derivations or code outputs with lexical variation). 
% To ensure comparability across datasets and models, all outputs are preprocessed using case normalization, punctuation removal, and whitespace standardization. 
The evaluation is performed using a unified reward interface with dataset-specific ground truth answers. The \textit{inference cost} is computed via a normalized pricing scheme that scales with the number of tokens and a model-specific base rate. The router’s average cost per dataset (Table~\ref{tab:cost_comparison}) reflects its actual per-query compute usage, offering a realistic picture of test-time efficiency.

\begin{table}[h]
\centering
\caption{Test quality score of \algo\, candidate small LLMs, and selected large LLMs across six benchmark datasets. \algo\ dynamically routes across models to optimize performance.}
\label{tab:acc_comparison}
\resizebox{\linewidth}{!}{%
\begin{tabular}{lcccccc}
\toprule
Model & MMLU & GSM8K & MBPP & MATH & ARC-Easy & ARC-Challenge \\
\midrule
\multicolumn{7}{l}{\textit{Bigger LLMs}} \\
Qwen2.5-14B-Instruct          & 0.003 & 0.077 & 0.028 & 0.076 & 0.003 & 0.003 \\ 
Llama-3.1-8B-Instruct         & 0.005 & 0.06 & 0.025 & 0.057 & 0.006 & {0.003} \\
\midrule
\multicolumn{7}{l}{\textit{Candidate LLMs}} \\
Qwen2.5-Coder-3B-Instruct     & 0.002 & {0.067} & {0.030} & 0.069 & 0.003 & {0.003} \\
DeepSeek-R1-Distill-Qwen-1.5B & 0.003 & 0.054 & 0.016 & 0.053 & 0.002 & 0.002 \\
Phi-3.5-mini-instruct         & {0.004} & {0.067} & \textbf{0.044} & 0.068 & 0.003 & {0.003} \\
\midrule
\algo                         & \textbf{0.005} & \textbf{0.139} & 0.029 & \textbf{0.105} & \textbf{0.009} & \textbf{0.010} \\
\bottomrule
\end{tabular}
}
\end{table} 

\paragraph{Quality Gains from Routing.} As seen in Table \ref{tab:acc_comparison}, \algo\ consistently outperforms all candidate small LLMs and often even the larger ones. On GSM8K, \algo\ achieves a quality score of \textbf{0.139}, more than doubling the best candidate baseline (Phi-3.5 at 0.067) and outperforming Llama-3.1-8B (0.060) and Qwen2.5-14B (0.077). On MATH, \algo\ yields a score of \textbf{0.105}, outperforming all small and large models, including Qwen2.5-14B (0.076). On MBPP, a code generation benchmark, \algo\ matches Qwen2.5-14B (0.028) and surpasses Llama-3.1-8B (0.025), while trailing the highest static score (Phi-3.5 at 0.044) by a modest margin. These results highlight how routing enables \algo\ to exploit complementarity across models without needing to re-train or ensemble them. The benefits of routing are especially clear in low-signal regimes like ARC-Challenge and ARC-Easy, where static models offer limited improvements beyond chance (e.g., scores around 0.002-0.003), \algo\ yields \textbf{0.010} and \textbf{0.009}, respectively, more than 3$\times$ better than the best-performing static candidate. These results highlight how multi-hop routing enables \algo\ to adapt to task difficulty and model strengths without requiring additional supervision or fine-tuning. Taken together, the results support our central hypothesis: dynamically routing queries over a diverse model pool yields higher quality responses than any fixed single-model strategy, including larger LLMs, while maintaining compute efficiency. \textbf{Comparison with bigger LLMs:} Table~\ref{tab:acc_comparison} also shows that even larger-scale models like Llama-3.1-8B and Qwen2.5-14B fall short of \algo\ on several datasets. For example, Llama-3.1-8B lags behind \algo\ by over 0.08 points on GSM8K and nearly 0.05 on MATH. % While these models offer stronger raw capacity, their effectiveness is inconsistent, and their average token-level inference cost (Table~\ref{tab:cost_comparison}) remains arguably similar or close to that of \algo. 

\begin{table}[h]
\centering
\caption{Average inference cost of \algo, candidate small LLMs, and large LLMs.} 
\label{tab:cost_comparison}
\resizebox{\linewidth}{!}{%
\begin{tabular}{lcccccc}
\toprule
Model & MMLU & GSM8K & MBPP & MATH & ARC-Easy & ARC-Challenge \\
\midrule
\multicolumn{7}{l}{\textit{Bigger LLMs}} \\
Qwen2.5-14B-Instruct  & 0.0095 & 0.0089 & 0.0092 & 0.0098 & 0.0086 & 0.0100 \\
Llama-3.1-8B-Instruct         & 0.0071 & 0.0078 & 0.0048 & 0.0077 & 0.0071 & 0.0096 \\
\midrule
\multicolumn{7}{l}{\textit{Candidate LLMs}} \\
Qwen2.5-Coder-3B-Instruct     & 0.0025 & 0.0030 & 0.0028 & 0.0031 & 0.0029 & 0.0030 \\
DeepSeek-R1-Distill-Qwen-1.5B & 0.0013 & 0.0015 & 0.0012 & 0.0016 & 0.0016 & 0.0017 \\
Phi-3.5-mini-instruct         & 0.0025 & 0.0027 & 0.0021 & 0.0029 & 0.0027 & 0.0030 \\
\midrule
\algo   & \textbf{0.0178} & \textbf{0.0163} & \textbf{0.0183} & \textbf{0.0159} & \textbf{0.0184} & \textbf{0.0188} \\
\bottomrule
\end{tabular}
}
\end{table} 

\begin{figure}[t]
    \centering
    \includegraphics[width=0.7\linewidth]{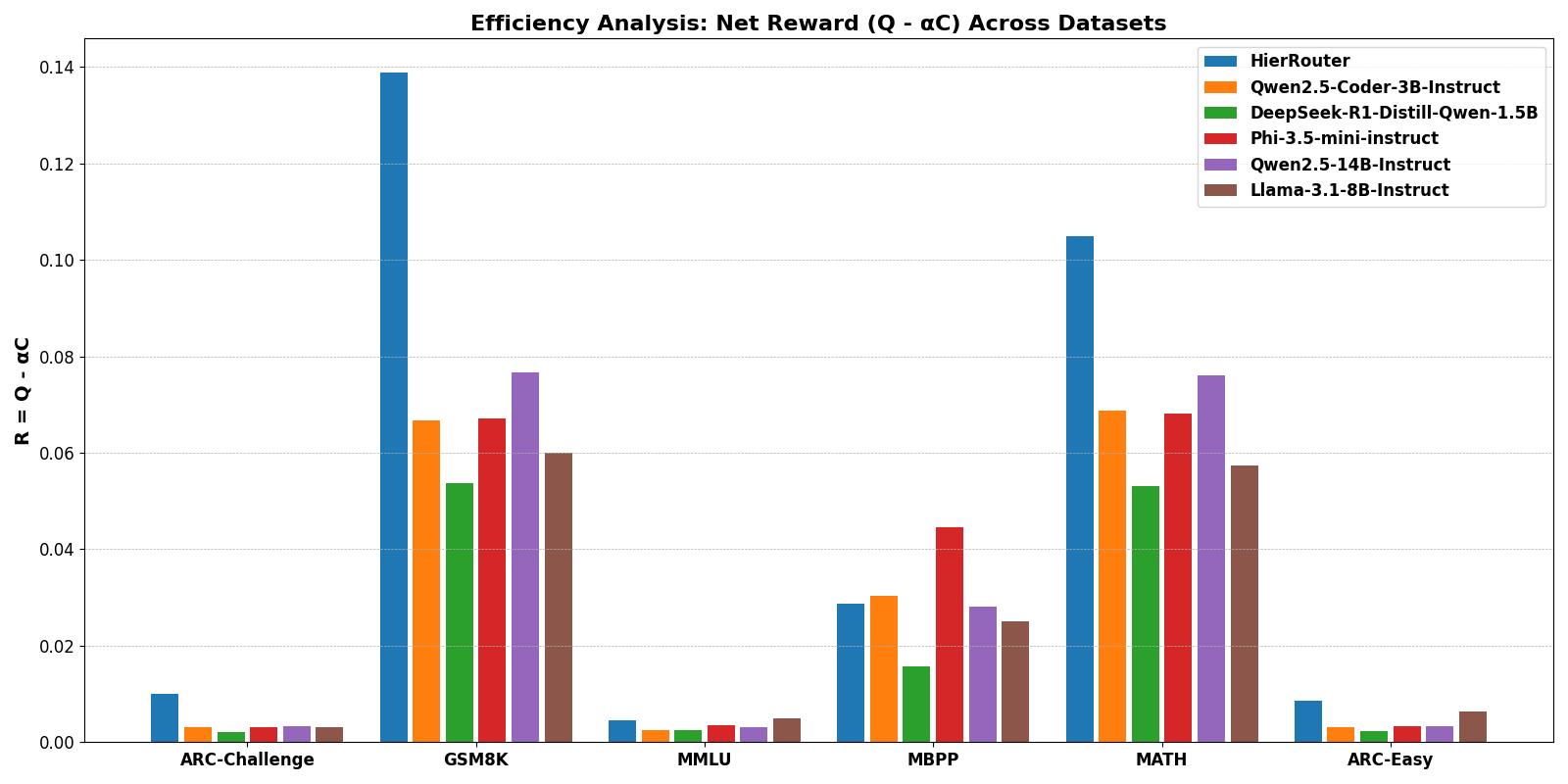}
    \caption{Comparing met score or reward comparison ($R = Q - \alpha C$). $Q$ denotes average F1 quality score, $C$ denotes average cost per query (token-based), and $\alpha$ is the cost penalty coefficient used during training. Despite slightly higher costs, \algo\ achieves superior or comparable net reward in all datasets, demonstrating efficient tradeoffs between performance and compute usage.} 
    \label{fig:rewards}
\end{figure}

\paragraph{Efficiency and Cost Analysis.} Each model is associated with a token-dependent inference cost, as defined earlier (in subsection \ref{sec:mdp_formulation}) via a normalized pricing formulation that scales with the number of input/output tokens and a model-specific base rate. This dynamic cost is integrated into the reward function used to train \algo, guiding it to balance output quality with inference efficiency. Table~\ref{tab:cost_comparison} presents the average token-normalized inference cost (per 1000 tokens) incurred by each model. As expected, the large models like Qwen2.5-14B and Llama-3.1-8B consistently incur higher compute costs (up to 0.01), while smaller candidates remain within the 0.001-0.003 range. In contrast, \algo\ incurs a moderate cost, between \textbf{0.0159} and \textbf{0.0188}, across all datasets. These values are higher than any single small model but substantially lower than invoking a very large LLM even once. This difference is a natural result of \algo's multi-hop routing framework. The router operates in two layers: the first generates initial decisions, and the second layer consumes enriched inputs that include prior outputs, embeddings, and query context. This increased input footprint leads to more tokens being processed overall, raising cost moderately. However, the benefit is clear: \algo's routing flexibility enables it to consistently deliver better-quality answers (Table \ref{tab:acc_comparison}), even outperforming the bigger models. To further quantify this tradeoff, Figure~\ref{fig:rewards} reports the net score or reward $R = Q - \alpha C$, combining quality score $Q$ and cost penalty $C$ using a training-time cost coefficient $\alpha$. Across nearly all tasks, \algo\ achieves the highest net reward, despite incurring slightly more cost, demonstrating that its dynamic policy makes efficient, context-sensitive use of compute. We argue that this tradeoff is both expected and desirable. The router pays a marginal premium for harder queries but avoids blanket overuse of high-cost models. In contrast, static baselines must commit to either efficiency (but low quality) or brute-force accuracy (with high, fixed cost). \algo\ sidesteps this tension with input-adaptive routing. % , delivering strong performance within practical budget constraints which is a valuable property in real-world applications. 

These results demonstrate the value of hierarchical, cost-aware routing. \algo's multi-hop policy architecture enables it to reason over both model specialization and budget constraints, leading to improved performance on a diverse set of tasks. Despite moderate increases in inference cost relative to small models, \algo\ achieves consistent quality gains, often exceeding the performance of individual candidates or bigger LLMs. This positions it as a scalable and practical alternative to brute-force LLM usage, making it particularly well-suited for deployment. % in compute-constrained applications requiring fine-grained control over inference efficiency. 

% \section{Discussion}
% \label{sec:Discussion}
% \input{sections/07_discussion}

\section{Conclusion}
\label{sec:Conclusion}

We introduced \algo, a hierarchical, RL-driven routing framework for composing specialized language models in a multi-hop inference pipeline. Unlike typical one-shot model invocation strategies, our method learns to make sequential, context-aware decisions that balance response quality with compute efficiency. By formulating the problem as an MDP and leveraging PPO for policy optimization, \algo\ effectively integrates task semantics and budget constraints into its routing logic. Extensive experiments across six diverse benchmarks demonstrate that \algo\ benefits from staged reasoning and model specialization and delivers substantial quality gains with minimal cost overhead. This paper affirms that RL-trained multi-hop routing is a promising alternative to brute-force scaling, enabling efficient and adaptive LLM deployment. Several future directions emerge. First, extending to dynamic model pools could enable lifelong adaptation to evolving domains. Second, integrating uncertainty-aware decision-making may enable robustness by deferring to stronger models under low confidence. Finally, combining routing with early-exit strategies could further amplify efficiency by stopping inference when adequate certainty is reached early. 

\bibliographystyle{plain} % We choose the "plain" reference style
\bibliography{refs} % Entries are in the refs.bib file 

\newpage
\section*{Appendix}
\label{sec:Appendix}
\noindent\rule{\linewidth}{1pt}

\noindent \textbf{Appendix A: Environment Dynamics.} Details the environment step function used to simulate multi-hop model invocation, reward calculation, and sequential decision-making.

\noindent \textbf{Appendix B: PPO and Router Configuration.} Covers training hyperparameters, architecture setup, routing prompts, and cost-based reward shaping.

\noindent \textbf{Appendix C: Evaluation.} Presents the token-level F1 score computation used to evaluate open-ended QA responses, including normalization and precision-recall logic.

\noindent\rule{\linewidth}{1pt}

\section*{Appendix A: Additional details on Environment Dynamics} 
\label{app:A}

This section outlines the custom environment step logic used to simulate hierarchical LLM routing as a sequential decision-making process. The design abstracts multi-hop routing as a finite-horizon MDP, where each decision selects a model and optionally terminates inference. The environment tracks the evolving query context, accumulates cost, and computes a final reward based on output quality and incurred cost.

\begin{algorithm}[h]
\caption{\algo\ Environment Step Function}
\label{alg:env-step}
\SetKwInOut{Input}{Input}
\SetKwInOut{Output}{Output}
\Input{Query $q$, action $a_t$ at time $t$, ground truth (optional)}
\Output{Next state $s_{t+1}$, reward $r_t$, done flag, info}

\BlankLine
\If{first step}{
    Initialize context $c_0 \gets q$, layer index $\ell_0 \gets 0$, total cost $C_0 \gets 0$\;
}

\If{multi-layer}{
    Parse action $a_t$ to get model index $m_t$ and stop flag $\texttt{done}_t$\;
}
\Else{
    Set $m_t \gets a_t$, $\texttt{done}_t \gets \texttt{True}$\;
}

\BlankLine
Obtain model response: $r_t \gets \texttt{Gen}(m_t, c_t)$\;
Compute cost: $\delta_t \gets \texttt{Cost}(m_t, c_t)$\;
Update total cost: $C_{t+1} \gets C_t + \delta_t$\;
Update context: $c_{t+1} \gets c_t \Vert r_t$\;
Increment layer: $\ell_{t+1} \gets \ell_t + 1$\;

\BlankLine
\If{$\texttt{done}_t$ or $\ell_{t+1} = L$}{
    Evaluate response quality: $q_t \gets Q(c_{t+1})$\;
    Compute reward: $r_t \gets q_t - \alpha C_{t+1}$\;
    Set $\texttt{done} \gets \texttt{True}$\;
}
\Else{
    Set $r_t \gets 0$, $\texttt{done} \gets \texttt{False}$\;
}

\BlankLine
Return state $s_{t+1} = (c_{t+1}, \ell_{t+1}, C_{t+1})$, reward $r_t$, done, and info\;

\end{algorithm}

\section*{Appendix B: PPO and Router Training Configuration}
\label{app:B}
This section details the configuration used to train \algo\ via Proximal Policy Optimization (PPO). The hierarchical router operates with two layers and a semantic encoder to guide multi-hop decision-making. The reward incorporates token-level quality (e.g., F1) and model-specific cost penalties.

\paragraph{PPO Training Configuration.}
The policy is trained using PPO with the following key parameters: 8 iterations, 128 rollouts per iteration, 16 mini-batches, 4 training epochs per iteration, $\gamma = 0.99$, GAE-$\lambda = 0.95$, clipping parameter 0.2, value loss coefficient 0.5, entropy coefficient 0.01, maximum gradient norm 0.3, and learning rate $10^{-4}$. The optimizer uses Adam with $\epsilon = 10^{-4}$. 

\paragraph{Router Architecture.}
The router comprises two decision layers. Inputs are embedded using a pre-trained encoder \texttt{sentence-transformers/all-mpnet-base-v2}\footnote{\url{https://huggingface.co/sentence-transformers/all-mpnet-base-v2}} with a maximum sequence length of 512. Each layer is prompted with task-specific instructions:

\begin{itemize}
    \item \textbf{Layer 0:} \textit{“Describe the problem in detail, then plan how you would solve it. Analyze the problem step by step, identifying key constraints and requirements.”}
    \item \textbf{Layer 1:} \textit{“Using the previous analysis and plan, verify if the approach is correct and solve the problem methodically. Ensure completeness and correctness in your solution.”}
\end{itemize}

\paragraph{Training Protocol.}
Training is performed on CUDA-enabled devices with a batch size of 96 for 200 epochs. A warm-up phase of 100 steps is used, with evaluations every 3 epochs and checkpoints every 5 epochs. A fixed seed of 42 ensures reproducibility.

\paragraph{Model Cost Configuration.}
Each LLM is associated with a normalized cost value used in computing the penalty term $c$:
\begin{center}
\begin{tabular}{lc}
\toprule
\textbf{Model} & \textbf{Cost} \\
\midrule

Qwen2.5-Coder-3B-Instruct     & 0.003 \\
DeepSeek-R1-Distill-Qwen-1.5B & 0.002 \\
Phi-3.5-mini-instruct         & 0.003 \\
Llama-3.1-8B-Instruct         & 0.008 \\
Qwen-14B-Instruct & 0.014 \\
\bottomrule
\end{tabular}
\end{center}
The reward function encourages economical routing while maintaining high answer quality.

\section*{Appendix C: Ground Truth Evaluator (F1 Score)}
\label{app:C}
To evaluate model responses against known correct answers in tasks such as QA and math reasoning, we implement a token-level F1 score evaluator. This serves as the primary external reward model during both training and evaluation.

\begin{algorithm}[h]
\caption{Evaluate(query, response, ground\_truth)}
\label{alg:f1_eval}
\Input{Query $q$, Model Response $r$, Ground Truth Answer(s) $g$}
\Output{F1 score $\in [0, 1]$}

% \begin{algorithm}[h]
% \caption{Evaluate(query, response, ground\_truth)}
% \label{alg:f1_eval}
% \Input{Query $q$, Model Response $r$, Ground Truth Answer(s) $g$}
% \Output{F1 score $\in [0, 1]$}
% Compute precision $p = \frac{|r \cap g|}{|r|}$\;
% Compute recall $r = \frac{|r \cap g|}{|g|}$\;
% Compute F1 score $f1 = \frac{2pr}{p + r}$\;
% \Return $f1$\;
% \end{algorithm} 

\BlankLine
Normalize $r$ and $g$: lowercase, strip punctuation, standardize whitespace\;

Tokenize $r \rightarrow r_{\text{tokens}}$, $g \rightarrow g_{\text{tokens}}$\;

Compute intersection: $I \gets r_{\text{tokens}} \cap g_{\text{tokens}}$\;

\If{$|r_{\text{tokens}}| = 0$ or $|g_{\text{tokens}}| = 0$ or $|I| = 0$}{
    \Return 0.0
}

\BlankLine
Compute precision: $p = \frac{|I|}{|r_{\text{tokens}}|}$\;

Compute recall: $r = \frac{|I|}{|g_{\text{tokens}}|}$\;

\Return $F1 = \frac{2 \cdot p \cdot r}{p + r}$

\end{algorithm}

\paragraph{Notes.} 
If multiple ground truth answers are available, we return the maximum F1 score among them. This formulation ensures robustness to lexical variation and partial correctness in free-form answers. The score serves as a reliable and differentiable reward signal for PPO optimization in our router framework. 

% \newpage
% \section*{NeurIPS Paper Checklist}
% \input{sections/checklist} 

\end{document}